\documentclass[10pt,twocolumn,letterpaper]{article}

\usepackage{cvpr}              % To produce the CAMERA-READY version
\makeatletter
\@namedef{ver@everyshi.sty}{}
\makeatother

\usepackage{graphicx}
\usepackage{amsmath}
\usepackage{amssymb}
\usepackage{booktabs}
\usepackage[normalem]{ulem}
\usepackage{soul}

\usepackage{todonotes}

\usepackage[breaklinks,colorlinks]{hyperref}

\usepackage[capitalize]{cleveref}
\crefname{section}{Sec.}{Secs.}
\Crefname{section}{Section}{Sections}
\Crefname{table}{Table}{Tables}
\crefname{table}{Tab.}{Tabs.}

\def\cvprPaperID{1} % *** Enter the CVPR Paper ID here
\def\confName{CVPR}
\def\confYear{2022}

\begin{document}

\title{Saccade Mechanisms for Image Classification, Object Detection and Tracking}

\author{
Saurabh Farkya\thanks{authors equally contributed to this work} \hspace{30pt} Zachary Daniels\footnotemark[1] \hspace{30pt} Aswin Nadamuni Raghavan \\
David Zhang \hspace{30pt} Michael Piacentino \\ \\
Center for Vision Technologies\\
SRI International\\
{\tt\small \{saurabh.farkya,zachary.daniels,aswin.raghavan,david.zhang,michael.piacentino\}@sri.com}
}
\maketitle

%%%%%%%%% ABSTRACT
\begin{abstract}
  
We examine how the saccade mechanism from biological vision can be used to make deep neural networks more efficient for classification and object detection problems.
Our proposed approach is based on the ideas of attention-driven visual processing and saccades, miniature eye movements influenced by attention. We conduct experiments by analyzing:
i) the robustness of different deep neural network (DNN) feature extractors to partially-sensed images for image classification and object detection, and
ii) the utility of saccades in masking image patches for image classification and object tracking. Experiments with convolutional nets (ResNet-18) and transformer-based models (ViT, DETR, TransTrack) are conducted on several datasets (CIFAR-10, DAVSOD, MSCOCO, and MOT17). Our experiments show intelligent data reduction via learning to mimic human saccades when used in conjunction with state-of-the-art DNNs for classification, detection, and tracking tasks. We observed minimal drop in performance for the classification and detection tasks while only using about 30\% of the original sensor data. We discuss how the saccade mechanism can inform hardware design via ``in-pixel'' processing. % \vspace{-20pt}
\end{abstract}

%\note{Might want to think of a better title}
%\note{Need to clean up a lot of the charts (\eg, unclear legends, missing comparisons, etc.)}
%\note{Saurabh: Where/how to add detection partial images?}
%\note{Need to maintain consistency in terminology; \eg gating vs filtering vs selecting}

%%%%%%%%% BODY TEXT
\section{Introduction}
\label{sec:intro}
\let\thefootnote\relax\footnotetext{The views, opinions and/or findings expressed are those of the author(s) and should not be interpreted as representing the official views or policies of the Department of Defense or the U.S. Government. \\ \textbf{DISTRIBUTION STATEMENT A.} Approved for Public Release,
Distribution Unlimited}Over the last decade, advances in Deep Neural Networks (DNN) have led to tremendous progress towards solving many computer vision tasks such as video-based scene understanding. In this paper, we consider the high resource use caused by the high-bandwidth data transmitted from sensor to DNN. The high data requirement puts increasingly powerful DNNs at odds with increasingly powerful sensors (e.g., increasing resolution in megapixels) in applications with cloud-based and edge device-based endpoints. We examine how mechanisms from biological vision can mitigate and reduce the data requirement to improve the practicability and efficiency of DNN-based end-to-end systems for image and video processing applications. Our approach is based on the ideas of attention-driven visual processing and saccades \cite{de1962essai}, miniature eye movements influenced by attention.
The focus of this paper is on understanding the effect of intelligent image patch selection as a pre-processing step for DNN-based image classification, object detection, and multi-object tracking.
First, we examine the robustness of state-of-the-art (SOTA) DNNs to random masking of image patches. We compare two broad types of DNNs: convolutional (ResNet-18) and self-attention-based transformers (ViT, DETR, TransTrack). Subsequently, we explore the benefits of intelligently filtering ``non-salient'' patches via saccades for classification and tracking. Experiments are conducted on several datasets: CIFAR-10, DAVSOD, MSCOCO, and MOT17. We observed minimal decreases in standard metrics while only using ${\sim}{30\%}$ of the original sensor data, and unlike recent work \cite{he2021masked}, our approach requires no fine-tuning of the SOTA DNN on partial images.

\section{Saccades as a Data-Reduction Mechanism}
%\note{Include an eye-fixation figure from DAVSOD @Saurabh}

Recently, there has been great interest in using attention mechanisms in computer vision models to improve model accuracy, improve model robustness to novel data, and reduce computation \cite{guo2022attention}.
These attention mechanisms are often inspired by human vision \cite{carrasco2011visual}: the human visual system has the capability of selecting important sensory information before passing the selected information to higher-level cortical areas where the brain performs further processing \cite{borji2013stands}.
In contrast to prior work, we examine the use of \textit{saccades}, which reduce data bandwidth by sending partially-sensed images (in our work by selecting subsets of image patches) to the DNN for processing. This is orthogonal to other methods for improving efficiency of DNNs such as network weight quantization.

%Attention mechanisms have been widely clustered into two categories: bottom-up attention and top-down attention \cite{katsuki2014bottom}. Bottom-up attention selects salient regions of sensory input based on inherent properties of the stimuli that make the regions stand out relative to the background. Top-down attention selects regions of the sensory input by exploiting known high-level context in the form of prior knowledge, plans, and goals. {\color{red} In this work, we consider a bottom-up attention mechanism that anticipates (filtering foreground patches from background patches for image inputs) and top-down attention (because the selection of patches at time $t+1$ are based on high-level understanding of a scene at time $t$).} \note{Need to check if this is an accurate statement.}

Saccades are quick and simultaneous eye movements between two or more phases of fixations influenced by attention. Saccadic movement of human eyes directs the fovea, the central part of the retina, by small increments (about 1-2 degrees) quickly to the point of interest. By moving the eye, small parts of a scene can be sensed with greater resolution \cite{provis2013adaptation}. Saccades are jerky motions that are discontinuous in directions and in magnitudes. However, the disconnected, fragmented scanning of the scene does not disrupt the feature integration \cite{reuther2020eye}. Studies have shown that saccadic motions can be compensated, and features are not mixed or misplaced. Furthermore, attention always shifts to the intended location before the eyes begin to move \cite{peterson2004covert}.

There is considerable evidence that ties attention to saccadic eye movements \cite{hoffman1995role}. Taking inspiration from the human visual system, we explore the effect of intelligent image patch selection as a means to reduce the amount of data that must be processed by DNNs, enabling the use of deep learning for real-time applications. 
This work can then lead to the design of new in-pixel or near-pixel computing sensors that only digitize and process minimal data as predicted from previous frames.

\noindent\textbf{Connection to Existing Work in Computer Vision}
Other work has attempted to efficiently explore portions of a scene and thus limit processing by some downstream model (\eg active vision \cite{aloimonos1988active,ballard1991animate,bajcsy1988active}). Recently, there has been interest in incorporating active dynamic perception with deep learning-based models, \eg Deep Anticipatory Networks \cite{satsangi2020maximizing}, Glance and Focus Networks \cite{huang2022glance}, shift-aware models for visual saliency object detection \cite{fan2019shifting}, and other deep reinforcement learning-based approaches for learning how to pan and zoom \cite{uzkent2020learning,uzkent2020efficient,caicedo2015active}. Furthermore, the use of attention and saliency are well-studied problems in the computer vision community \cite{nguyen2018attentive}. Similar to our findings, recent work \cite{he2021masked,bachmann2022multimae} has shown that if properly trained, transformers (DNNs based on self-attention) can successfully capture the image context even with extreme masking. There have also been past approaches that leverage saccades as a means of improving DNNs, \eg, Recurrent Attention Models, which first sense a coarse/blurry image is sensed completely, and then saccades refine certain image locations \cite{mnih2014recurrent,ba2014multiple}. Our approach differs from others because (a) we aim to understand whether mimicking human saccades (using patch-based processing) is an effective data reduction mechanism, and (b) whether the performance of SOTA DNNs be maintained without finetuning to partially sensed images. \vspace{-10pt}

% \section{Understanding the Affects of Gating for Image Classification}
\section{Saccades for Image Classification, Object Detection and Tracking}
% \vspace{-10pt}
In this section, we show the effect of partially obscuring images before passing through a DNN image classifier. We aim to validate the following hypotheses: i) Self-attention-based transformers \cite{dosovitskiy2020image} are robust to partially-sensed patch-based inputs, and ii) Mimicking saccadic motions captured via human eye-fixation data reduces the amount of information that must be processed by deep learning models while sacrificing minimal amounts of classification accuracy.  

%\note{@David: Justification for using Transformer models is required} 

%

\subsection{Transformer-Based Models are Robust to Partially-Sensed Patch-Based Inputs}
\begin{figure}[t]
  \centering
    % \vspace{-10pt}
    \includegraphics[width=0.95\linewidth]{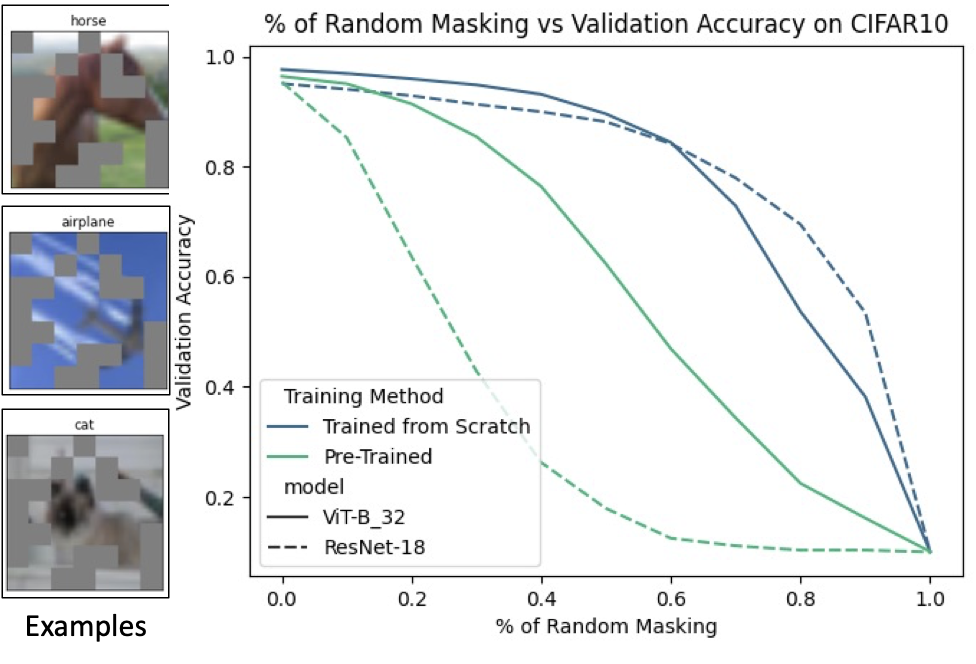}
    \caption{(Left) Example CIFAR10 images with random patches masked. (Right) Image classification accuracy of convolutional vs vision transformer with increasing masking (ResNet-18 vs ViT). When models are trained from scratch, a model is learned for each level of masking.}
    \label{fig:cifar_random}
\end{figure}

%\note{TODO: Transformers are useful from hardware-perspective; @Saurabh: add sentence on artifacting in CNNs vs Transformers}
We look at the effect of \textit{random} patch selection on two classes of DNN architectures for image classification: convolutional neural networks (CNNs) viz.\! ResNet-18 \cite{he2016deep} and a transformer-based model viz.\! ``Vision Transformer (ViT)'' \cite{dosovitskiy2020image}.
We hypothesize that the vision transformer, which breaks images into patches and processes the rasterized sequence of patches using self-attention, will be more robust to random masking of patches in the input image (see Fig.~\ref{fig:cifar_random} (left) for examples). This is inline with recent work \cite{he2021masked,bachmann2022multimae}, which shows that transformers fine-tuned with heavy input masking are capable of reconstruction of the missing patches. Transformers are ideal for input patch masking because they naturally reduce processing of masked patches and are compatible with new hardware designs (Section \ref{sec:discussion}). We conduct our first experiment on the CIFAR-10 dataset \cite{krizhevsky2009learning} where we vary the percentage of masked patches and see the effect on various architectures in Figure~\ref{fig:cifar_random}.

%\note{@Saurabh This needs to be cleaned up; either include the after training for resnet or remove it for Txs}

The transformer (ViT) is significantly more robust to random masking of patches when using a pre-trained model. Fine-tuning ViT improves further, maintaining above 80\% accuracy even with only 40\% of the image sensed. In comparison, the pretrained CNN is significantly less robust with a faster drop in accuracy. Fine-tuning improves accuracy for both models; however, it could be a result of overfitting. % and the robustness is unclear across masking levels. % , suggesting that transformer-based models do not need to process entire images to achieve good performance on image classification tasks.

\begin{figure}[t]
  \centering
  \includegraphics[width=0.9\linewidth]{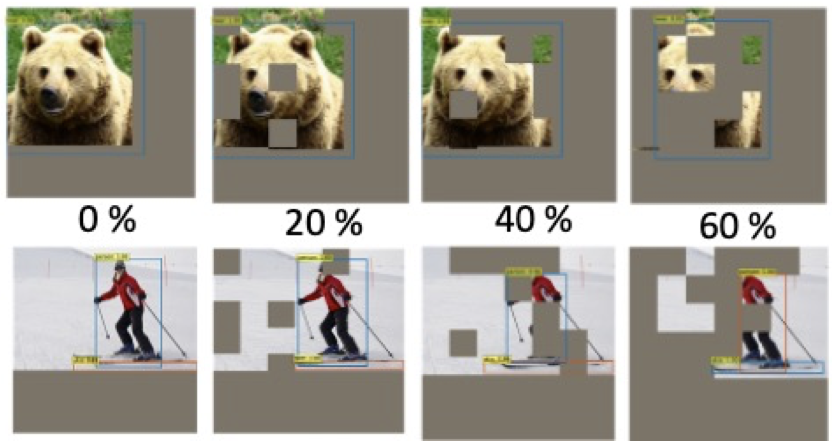}
    \caption{Example DETR detections at various masking levels.}
    \label{fig:coco_detections}
\end{figure}

\begin{figure}[t]
  \centering
  \includegraphics[width=0.95\linewidth]{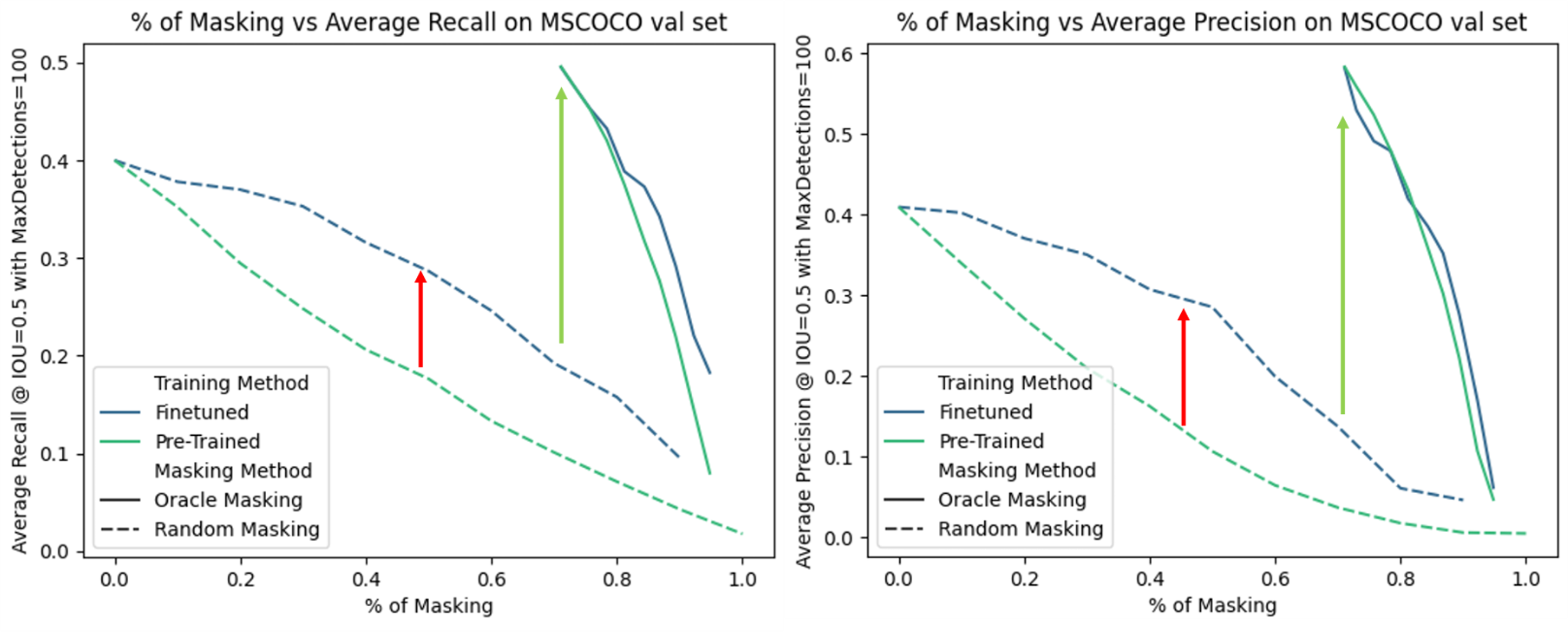}
    \caption{Robustness of the DETR detector to masked images from COCO: random vs oracle patch selection; mAP (left) and mAR (right) metrics at different levels of masking}
    \label{fig:coco_results}
\end{figure}

Next, we examine the effect of random patch selection for object detection. We subsample  the MSCOCO object detection benchmark \cite{lin2014microsoft} for manageable experimentation. The subsampled set contains 18,403 training and 800 validation images taken from the MSCOCO training and test sets, respectively$^1$. %, with 5 classes: car, bus, truck, traffic light and stop sign.
\footnote[2]{$^1$Subsampled MSCOCO available  \href{https://drive.google.com/drive/folders/1Vp-nZOVgWlOg3hERTZ-X7W14ALSmMSTB?usp=sharing}{here}.} 
We use DETR \cite{carion2020end} as the transformer-based object detector. We measure the mean average precision (mAP) and mean average recall (mAR) at different levels of masking and consider (i) random selection of patches and, (ii) foreground patches selected from instance segmentation ground truth. 
Qualitative results appear in Figure~\ref{fig:coco_detections} where we observe that objects are correctly detected despite missing one or more patches in the interior of the bounding box (when selecting patches intelligently). Quantitative results (Figure~\ref{fig:coco_results}) suggest that the DETR model significantly benefits from selecting only foreground patches (green arrow). Gains in mAP and mAR are also seen when finetuning (red arrow) but improvement from selecting foreground patches is noticeably larger and finetuning does not appear to be necessary in the latter case.

%\note{@Saurabh: Plot needs to updated with cleaner legends}

\begin{figure}[t]
  \centering
    \includegraphics[width=0.65\linewidth]{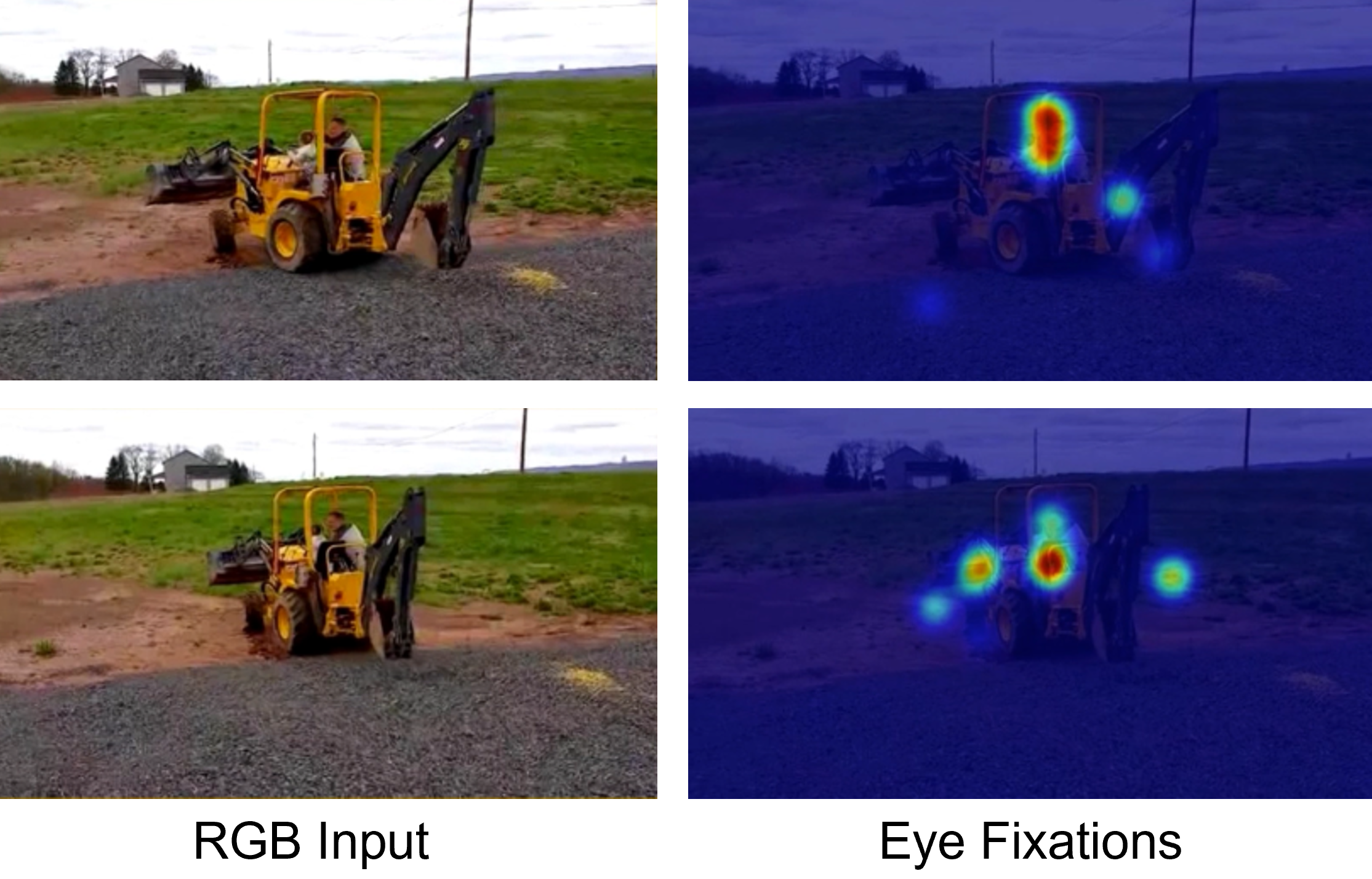}
    \caption{Example from the DAVSOD dataset \cite{fan2019shifting} where the human focuses on different objects as time progresses.}
    \label{fig:saliency_shift}
\end{figure}

\begin{figure}[t]
  \centering
    \includegraphics[width=0.8\linewidth]{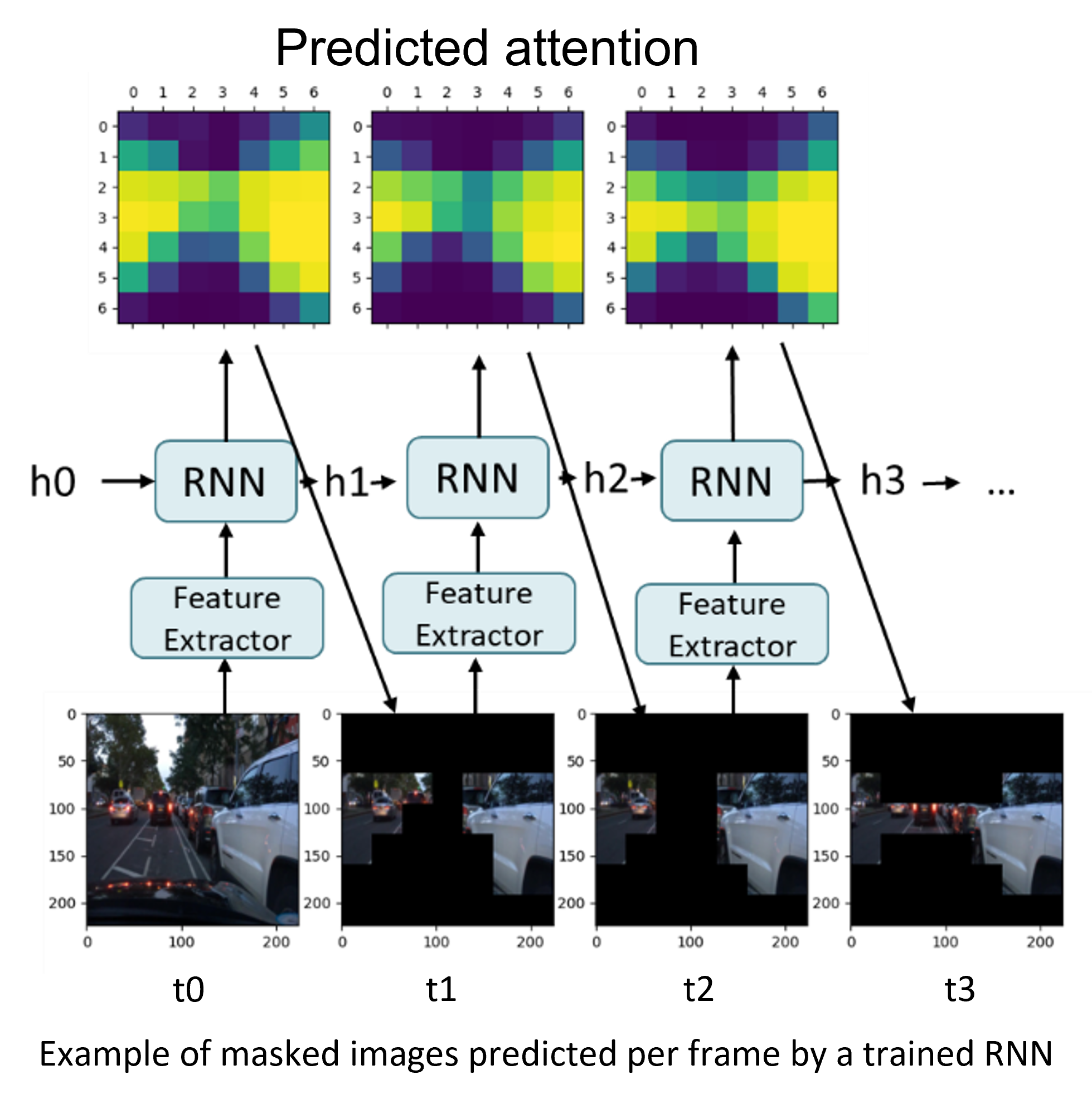}
    % \caption{Left: Example from the DAVSOD dataset \cite{fan2019shifting} where the human focuses on different objects as time progresses. Right: Training schema to mimic the  attention shift using an RNN.}
    \caption{Training schema to mimic attention shift using an RNN.}
    \label{fig:rnn}
\end{figure}

\subsection{Learning Saccades}
% \note{Need to emphasize that this is for video data and not applicable for CIFAR10 and COCO- maybe in the title of the section too}
% Next, we examine the trade-off between performance and data reduction for different patch selection mechanisms.
Next, we consider video data (unlike previous experiments). 
We train a recurrent DNN (an RNN) to anticipate human attention, thus mimicking saccades.
The DAVSOD dataset \cite{fan2019shifting} contains videos annotated with human eye fixation data. 
This dataset provides saliency shift ground truth where humans rapidly attend to different objects in a dynamically-changing scene (see Figure~\ref{fig:saliency_shift}).
We look at the effect of selecting patches that contain salient foreground objects based on the human eye fixations at time $t$.
We compared three methods: i) select random patches, ii) select patches that overlap with salient objects (objects that contain at least one fixation), and iii) select patches predicted by a trained recurrent neural network (RNN) \cite{cho2014properties}.
The RNN is trained to predict the sequence of  human attention at future time steps based on partially sensed inputs at earlier time steps.
Figure~\ref{fig:rnn} shows the training method for the RNN to predict human attention at times $t$+1, $t$+2, $t$+3. During training, every fourth video frame is fully-sensed after the RNN state is reset to $h_0$. During testing, only the first frame of the test video is fully-sensed, and the RNN state only resets between videos. The evaluation is per-frame image classification accuracy on four classes (human, animal, artifact, and vehicle) from the DAVSOD dataset.
The training optimizes binary cross-entropy with ground truth attention masks. At each time step, the RNN outputs a heatmap from which the top-$k$ patches are selected ($k$ is varied in the experiment).  

\begin{figure}[t]
  \centering
    \includegraphics[width=0.85\linewidth]{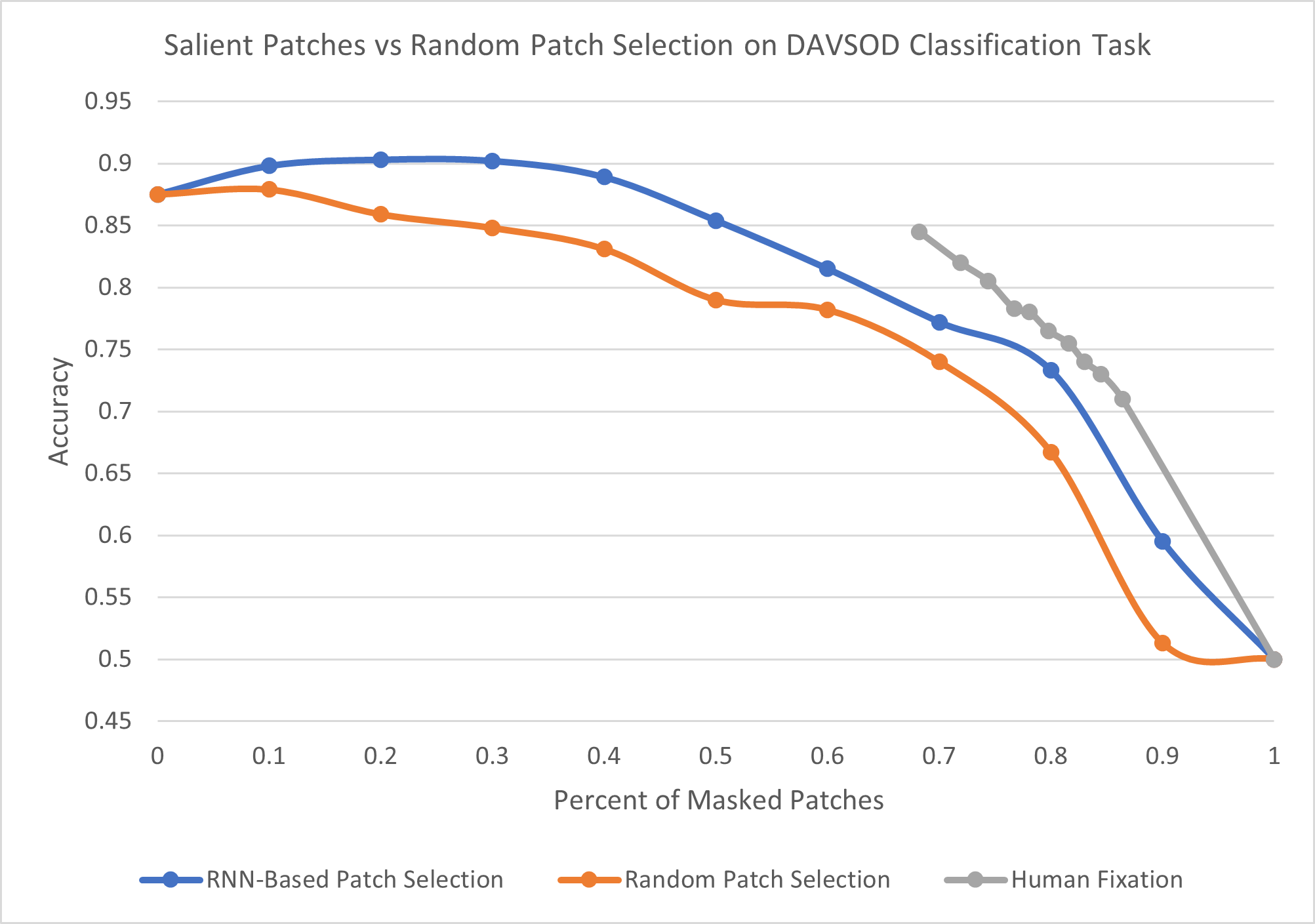}\\
    \includegraphics[width=0.85\linewidth]{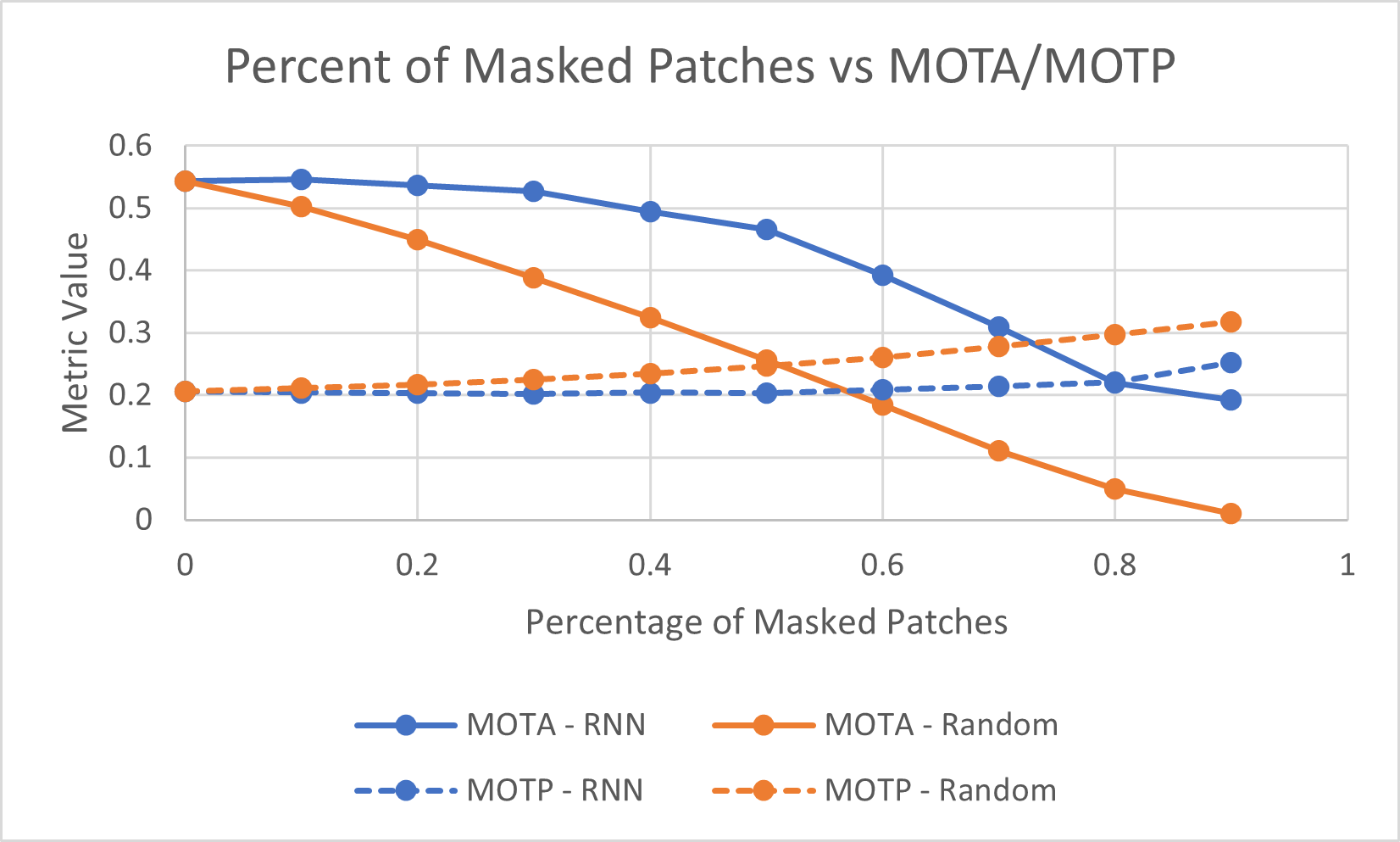}
    \caption{Top: Comparison of learned saccades vs human-driven saccade on the DAVSOD dataset. Note that the human fixation data points are determined by thresholding at different percentages of salient pixels per patch. Bottom: Comparing the RNN-based saccade model with random patch selection for multi-object tracking on MOT-17}
    \label{fig:davsod_classification}
\end{figure}

Our experiments suggest that reducing data via mimicking human saccades is noticeably more effective (higher accuracy) than random selection of patches. In Figure~\ref{fig:davsod_classification} (top), accuracy is not degraded when at least 30\% patches are selected using the oracle (gray line). Our RNN is able to reasonably predict where humans will look (AUROC of $0.78$ when trained to select 30\% of patches). In Figure~\ref{fig:davsod_classification} (top, blue line), the RNN-based patch selection noticeably outperforms random patch selection (orange line) and only slightly under-performs true human attention (gray line).

Our final experiment is on multi-object tracking where saccades can be useful for capturing object motion. We applied our learned RNN-based saccade to a pre-trained transformer-based tracker viz.\! TransTrack \cite{sun2020transtrack}.
We compared our RNN to random patch selection on the MOT17 pedestrian tracking benchmark$^2$ \cite{milan2016mot16}.
\footnote[3]{$^2$Training and validation videos were redistributed so training and validation videos were disjoint, see \href{https://drive.google.com/drive/folders/1VwK9Wp_5fW3447bLhmHlXd7jEAc_jgTQ?usp=sharing}{here for data and example outputs}.} This dataset does not provide human attention, so we used foreground object location maps as a proxy for training the RNN.
We evaluated using the MOTA and MOTP metrics \cite{milan2016mot16} and found noticeable improvement in both metrics (Figure~\ref{fig:davsod_classification} (bottom)) when using the learned saccades over the random patch selection in terms of both MOTA (higher better) and MOTP (lower better). See here\footnotemark[2] for example tracking results.

% \begin{figure}[t]
%   \centering
%     \includegraphics[width=0.92\linewidth]{graphics/tracking_metrics.png}
%     \caption{Comparing the RNN-based saccade model with random patch selection for multi-object tracking on the MOT-17 dataset}
%     \label{fig:mot_results}
%     \vspace{-8pt}
% \end{figure}

\section{Discussions and Future Directions}
\label{sec:discussion}

We examined how saccades can improve the efficiency of deep learning via reducing data processing. We conducted a thorough set of experiments aimed at analyzing i) the robustness of transformer-based DNNs to sensing partial images for ``coarse'' tasks like image classification and ``fine-grained'' tasks like object detection and tracking and ii) the utility of learned saccades-based selection of patches for use with deep classifiers and multi-object trackers.

\textbf{Hardware Benefit: } The saccade mechanism provides a new way for more efficiently processing pixels at the hardware-level. Video generates large amounts of data due to its high resolution, high dynamic range, and high frame rates. Thus, AI processing of video presents unique challenges for sensing at the edge. Processing in the sensor pixel (``in-pixel processing'') before data is moved to a traditional compute device is an effective way to reduce power and bandwidth at the sensor edge. Saccades are compatible with this paradigm by reducing non-salient data, potentially reducing the data bandwidth and feature dimensionality by an order of magnitude before the captured data is sent to the processing unit behind the imager. A patch that is not sensed does not need to be read-out from the analog pixel (ROIC) and digitized (ADC) leading to savings in power. 
% ${>}10{\times}$ before the captured data is sent to the processing unit behind the imager. 

% The sensor captures data and converts photons to electrons per pixel. The back-end processor predicts the salient patches for the next frame based on the history of past frames, and the analog circuit only processes these sparse salient patches into semantic features, which are converted to digital signals by ADCs and processed by near-pixel processing. The saccade-based in-pixel processing provides at least one magnitude of savings in terms of energy and latency without degrading performance.

\textbf{Future Work:} There is still work to be done in this area. We assumed we had access to human eye fixations or proxy ground truth information to guide the saccade mechanism. In many cases, we discarded background patches, which can provide useful information for processing novel scenes. We expect the model could be improved by directly learning from interaction which patches are and are not  informative. 
% Similarly, we only considered tasks where movement does not play a large role. In the future, we want to look at how saccades can be applied to problems requiring knowledge of object dynamics such as tracking. 
Such saccades could account for the confidence of predictions of individual objects, and explicitly tackle the tradeoff between exploitation to increase confidence with exploration to find new objects using Reinforcement Learning. 
%Finally, our models still processed the masked patches despite containing only zeros, resulting in unnecessary computation, which can be avoided in the future. 
% . We think it is important to extend our transformer models to process only the selected patches.

%\cite{munoz2004look}

% \begin{figure*}
%   \centering
%   \begin{subfigure}{0.68\linewidth}
%     \fbox{\rule{0pt}{2in} \rule{.9\linewidth}{0pt}}
%     \caption{An example of a subfigure.}
%     \label{fig:short-a}
%   \end{subfigure}
%   \hfill
%   \begin{subfigure}{0.28\linewidth}
%     \fbox{\rule{0pt}{2in} \rule{.9\linewidth}{0pt}}
%     \caption{Another example of a subfigure.}
%     \label{fig:short-b}
%   \end{subfigure}
%   \caption{Example of a short caption, which should be centered.}
%   \label{fig:short}
% \end{figure*}

% \begin{table}
%   \centering
%   \begin{tabular}{@{}lc@{}}
%     \toprule
%     Method & Frobnability \\
%     \midrule
%     Theirs & Frumpy \\
%     Yours & Frobbly \\
%     Ours & Makes one's heart Frob\\
%     \bottomrule
%   \end{tabular}
%   \caption{Results.   Ours is better.}
%   \label{tab:example}
% \end{table}

% When placing figures in \LaTeX, it's almost always best to use \verb+\includegraphics+, and to specify the figure width as a multiple of the line width as in the example below
% {\small\begin{verbatim}
%   \usepackage{graphicx} ...
%   \includegraphics[width=0.8\linewidth]
%                   {myfile.pdf}
% \end{verbatim}
% }

\section*{Acknowledgements}
We thank DARPA and Dr.~Mason, program manager for DARPA’s AIE IP2 program, for giving SRI the opportunity to explore and present new concepts in in-pixel computing. \vspace{-5pt}

%%%%%%%%% REFERENCES
{\scriptsize
\bibliographystyle{ieee_fullname}
\bibliography{PaperForReview}

\begin{thebibliography}{10}\itemsep=-4pt

\bibitem{aloimonos1988active}
J. Aloimonos et~al.
\newblock Active vision.
\newblock {\em IJCV}, 1988.

\bibitem{ba2014multiple}
J. Ba et~al.
\newblock Multiple object recognition with visual attention.
\newblock {\em arXiv}, 2014.

\bibitem{bachmann2022multimae}
R. Bachmann et~al.
\newblock Multimae: Multi-modal multi-task masked autoencoders.
\newblock {\em arXiv}, 2022.

\bibitem{bajcsy1988active}
R. Bajcsy.
\newblock Active perception.
\newblock {\em Proceedings of IEEE}.

\bibitem{ballard1991animate}
D. Ballard.
\newblock Animate vision.
\newblock {\em Artificial intelligence}, 1991.

\bibitem{borji2013stands}
A. Borji et~al.
\newblock What stands out in a scene? a study of human explicit saliency
  judgment.
\newblock {\em Vision research}, 2013.

\bibitem{caicedo2015active}
J.~C. Caicedo et~al.
\newblock Active object localization with deep reinforcement learning.
\newblock In {\em ICCV}, 2015.

\bibitem{carion2020end}
N. Carion et~al.
\newblock End-to-end object detection with transformers.
\newblock In {\em ECCV}, 2020.

\bibitem{carrasco2011visual}
M. Carrasco.
\newblock Visual attention: The past 25 years.
\newblock {\em Vision research}, 2011.

\bibitem{cho2014properties}
K. Cho et~al.
\newblock On the properties of neural machine translation: Encoder-decoder
  approaches.
\newblock {\em arXiv}, 2014.

\bibitem{de1962essai}
H. de Barjac et~al.
\newblock Essai de classification biochimique et s{\'e}rologique de 24 souches
  debacillus du typeb. thuringiensis.
\newblock {\em Entomophaga}, 1962.

\bibitem{dosovitskiy2020image}
A. Dosovitskiy et~al.
\newblock An image is worth 16x16 words: Transformers for image recognition at
  scale.
\newblock {\em arXiv}, 2020.

\bibitem{fan2019shifting}
D.-P. Fan et~al.
\newblock Shifting more attention to video salient object detection.
\newblock In {\em CVPR}, 2019.

\bibitem{guo2022attention}
M.-H. Guo et~al.
\newblock Attention mechanisms in computer vision: A survey.
\newblock {\em Computational Visual Media}, 2022.

\bibitem{he2016deep}
K. He et~al.
\newblock Deep residual learning for image recognition.
\newblock In {\em CVPR}, 2016.

\bibitem{he2021masked}
K. He et~al.
\newblock Masked autoencoders are scalable vision learners.
\newblock {\em arXiv}, 2021.

\bibitem{hoffman1995role}
J.~E. Hoffman et~al.
\newblock The role of visual attention in saccadic eye movements.
\newblock {\em Perception \& psychophysics}, 1995.

\bibitem{huang2022glance}
G. Huang et~al.
\newblock Glance and focus networks for dynamic visual recognition.
\newblock {\em arXiv}, 2022.

\bibitem{krizhevsky2009learning}
A. Krizhevsky et~al.
\newblock Learning multiple layers of features from tiny images.
\newblock 2009.

\bibitem{lin2014microsoft}
T.-Y. Lin et~al.
\newblock Microsoft coco: Common objects in context.
\newblock In {\em ECCV}, 2014.

\bibitem{milan2016mot16}
A. Milan, L. Leal-Taix{\'e}, I. Reid, S. Roth, and K. Schindler.
\newblock Mot16: A benchmark for multi-object tracking.
\newblock {\em arXiv preprint arXiv:1603.00831}, 2016.

\bibitem{mnih2014recurrent}
V. Mnih et~al.
\newblock Recurrent models of visual attention.
\newblock {\em NeurIPS}, 2014.

\bibitem{nguyen2018attentive}
T. Nguyen et~al.
\newblock Attentive systems: A survey.
\newblock {\em IJCV}, 2018.

\bibitem{peterson2004covert}
M. Peterson et~al.
\newblock Covert shifts of attention precede involuntary eye movements.
\newblock {\em Perception \& psychophysics}.

\bibitem{provis2013adaptation}
J.~M. Provis et~al.
\newblock Adaptation of the central retina for high acuity vision: cones, the
  fovea and the avascular zone.
\newblock {\em Progress in retinal and eye research}, 2013.

\bibitem{reuther2020eye}
J. Reuther et~al.
\newblock The eye that binds: Feature integration is not disrupted by saccadic
  eye movements.
\newblock {\em Attention, Perception, \& Psychophysics}, 2020.

\bibitem{satsangi2020maximizing}
Y. Satsangi et~al.
\newblock Maximizing information gain in partially observable environments via
  prediction rewards.
\newblock In {\em AAMAS}, 2020.

\bibitem{sun2020transtrack}
P. Sun et~al.
\newblock Transtrack: Multiple object tracking with transformer.
\newblock {\em arXiv}, 2020.

\bibitem{uzkent2020efficient}
B. Uzkent et~al.
\newblock Efficient object detection in large images using deep reinforcement
  learning.
\newblock In {\em WACV}, 2020.

\bibitem{uzkent2020learning}
B. Uzkent et~al.
\newblock Learning when and where to zoom with deep reinforcement learning.
\newblock In {\em CVPR}, 2020.

\end{thebibliography}
}

\end{document}